\documentclass{article} 
\usepackage{iclr2021_conference,times}
\iclrfinalcopy

\usepackage{amsmath,amsfonts,bm}

\def\eqref#1{equation~\ref{#1}}

\def\1{\bm{1}}

\DeclareMathAlphabet{\mathsfit}{\encodingdefault}{\sfdefault}{m}{sl}
\SetMathAlphabet{\mathsfit}{bold}{\encodingdefault}{\sfdefault}{bx}{n}

\usepackage{hyperref}
\usepackage{url}
\usepackage{graphicx}
\usepackage{enumitem}

\title{Visualizing Neural Network Imagination}

\author{Nevan Wichers \\
Google \\
\and
\textbf{Victor Tao} \\
University of Toronto \\
\and
\textbf{Riccardo Volpato} \\
Satalia \\
\and  
\textbf{Fazl Barez} \\
University of Oxford
}

\begin{document}

\setlist{nolistsep}

\maketitle

\begin{abstract}
In certain situations, neural networks will represent environment states in their hidden activations. Our goal is to visualize what environment states the networks are representing. We experiment with a recurrent neural network (RNN) architecture with a decoder network at the end. After training, we apply the decoder to the intermediate representations of the network to visualize what they represent. We define a quantitative interpretability metric and use it to demonstrate that hidden states can be highly interpretable on a simple task. We also develop autoencoder and adversarial techniques and show that benefit interpretability.
\end{abstract}

\section{Introduction}\label{intro}

When trained to predict the end state of a given sequence, neural networks (NNs) may learn to predict intermediate states even if not explicitly trained to do so. For example, a network trained to predict the end state of an environment where the easiest way is to predict it one step at a time will most likely generate hidden representations of the intermediate steps. Consider a NN trained to predict the 4th state of Conway's Game of Life (GoL) \citep{conway1970game} given an initial state. GoL is a cellular automata with simple deterministic rules to obtain the next state from the previous. The network could learn the GoL rules and apply them to the 1st state to obtain a representation of the 2nd state, to the 2nd state to obtain the 3rd state and so on. The states before the 4th state would be represented only internally in the network, since it is only trained to output the 4th state. Another case where the network might represent environment states is when making a decision. For example, if a reinforcement learning (RL) agent needs to decide whether to go left or right in a maze, the NN might predict what the state of the environment will be after going left. The network may represent this state in its hidden activations and use it to decide how to turn. Our goal is to visualize the environment states that a network is representing. In this case, it is like the network is imagining the consequence of its action. When a network is making a decision, visualizing the states it is considering can help evaluate if the network is making a safe and correct decision and for the right reasons. To our knowledge, this interpretability problem has not been explored before.

In this paper we focus on visualizing the network's intermediate representations when predicting GoL states. We use an RNN architecture with an encoder in the beginning and a decoder at the end. We train the network to predict a future GoL state given the initial GoL state. After training, we run inference with a hidden state of the RNN fed into the decoder. We hypothesize that the decoder will produce an image similar to an intermediate GoL state. 
We use the GoL because we know the ground truth of the intermediate GoL states so we can accurately evaluate our technique. For more interesting problems we cannot know what intermediate states the network is representing in advance. 

\subsection{Contributions}
\begin{enumerate}
    \item We introduce the task of visualizing the intermediate environment states represented by a neural network.
    \item We develop an adversarial training and autoencoder method for visualizing intermediate representations.
    \item We present results to show how the network architecture and training decisions impact interpretability.
    \item We show that the network often learns to predict intermediate states earlier than it learns to predict the state it is trained to.
\end{enumerate}


\section{Methodology}
In the following sections, we describe how our method applies to the GoL for clarity. However, our method is general-purpose and applicable to other domains like the maze example we described.
\subsection{Architecture and training}
\begin{figure}[!ht]
  \includegraphics[width=1.0\columnwidth]{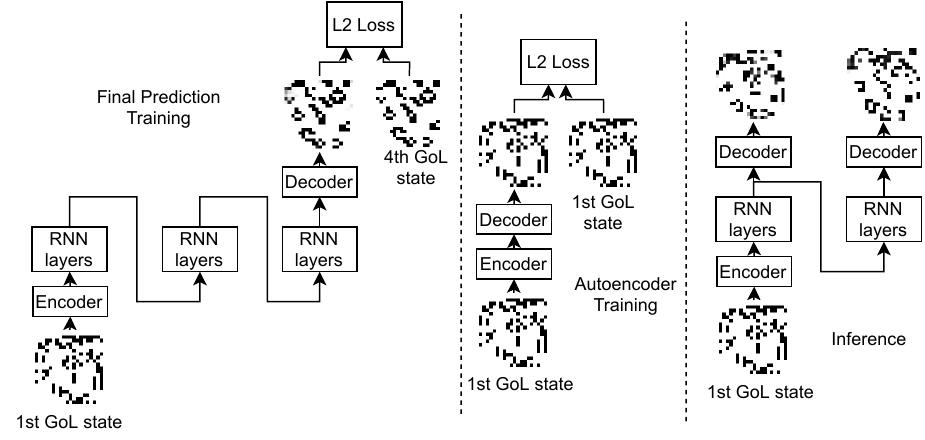} 
  \caption{\label{fig:setup_diagram} Our setup with 3 GoL and 3 model timesteps. \textbf{Left}: Our setup during training. The RNN layer blocks share weights. The Encoder, RNN layers and Decoder are all convolutional. \textbf{Middle}: Our autoencoder training setup. \textbf{Right}: Our setup during inference. Our hypothesis is that the states that are generated when applying the decoder to the intermediate timesteps are similar to the intermediate GoL states.}
\end{figure}

The left side of figure \ref{fig:setup_diagram} shows our architecture with 3 RNN timesteps and 3 GoL timesteps. By timestep we mean a transition from one state to the next in the GoL or the RNN. We also experiment with different numbers of RNN and GoL timesteps. Note that the network is only trained to predict the final game of life state, not the intermediate ones. We generate the first GoL state randomly\footnote{We randomly generate the 1s and 0s of the state. Sometimes we use the rules of GoL to generate the state to produce interesting patterns in the first state.}. Then we use the rules of GoL to calculate the $k$-th state. We use pairs of the first GoL state and the $k$-th state as training data. Each state is an image with 1 color channel where each pixel is either 0 or 1.

\subsection{Autoencoder training}
The encoder and decoder are also trained together like an autoencoder as shown in the middle of figure \ref{fig:setup_diagram}. We use L2 loss to train the combined encoder decoder to recreate the 1st GoL state. The autoencoder loss is averaged with the regular training loss. 
Our motivation is to make the encoder use a representation which the decoder can decode. We hope that this will encourage the intermediate RNN layers to also use a representation which the decoder can decode into a valid intermediate state. 

\subsection{Adversarial decoder training}
We also try to make the inferred intermediate look like GoL states by training the decoder like a generator in a Generative Adversarial Network (GAN) \citep{goodfellow2014generative}. Specifically, we train a discriminator to differentiate between real GoL states and states generated by the decoder. At the same time, we train the decoder to fool the discriminator. We apply this method with a pretrained RNN and encoder. We use the GAN loss, in addition to the autoencoder and final prediction losses to update only the weights of the decoder, leaving the encoder and RNN unaffected. Note that this method does not train the decoder to output a correct intermediate GoL state, only something that looks like a GoL state.

\subsection{Inference with decoder}
During training, the output of the intermediate RNN timesteps are only fed into the next RNN timestep. During inference, the output of the intermediate RNN timesteps are also fed into the decoder as shown on the right of figure \ref{fig:setup_diagram}. We hypothesize that applying the decoder to the intermediate states like this will generate the intermediate GoL states. We do not run inference on the output from the encoder or the final RNN timestep because these are already trained to match respectively the first and last GoL states.

\section{Results}\label{results}
\subsection{Metric}
We define a metric to determine how closely the network represents the intermediate GoL states. We consider a GoL sequence with $m$ states $\{s_{1} \ldots s_{m}\}$ and feed $s_{1}$ into the inference network to generate $\{h_{1} \ldots h_{n}\}$ outputs from the $n$ hidden states of the network. We threshold each pixel of network outputs at .5 so they become either 0 or 1, since GoL states have binary pixel values. 
Finally, we count the number of $h_i$ outputs that match any $s_j$ GoL state and divide it by the minimum between $n$ and $m-2$ (not $m$ since we exclude $s_1$ and $s_m$) as shown in this equation:
\begin{equation}
\frac{\sum_{i=1}^n \sum_{j=2}^{m-1}  f(h_i, s_j)}{\min(n, m-2)} \\
\end{equation}
$f$ is a function which determines if two states match. If less than 95\% of the pixels in the two inputs of $f(h_i, s_j)$ are equal $f$ gives 0. Otherwise, $f$ gives .5 if another generated state already matched the GoL state, and 1 if it did not. This gives credit for states in any order matching, because when the number of model and GoL timesteps are different it is not clear which states should match. Note that the calculated metric could be greater than 1 if $n > m - 2$. Each metric value that we report is the average over ten thousand GoL sequences for each run. See appendix \ref{appen:results-samples} for examples of metric values of state sequences.


\subsection{Results with same number of model and GoL timesteps}
First, we evaluate our technique with the same number of GoL and model timesteps. For each run, we try our technique with and without adversarial training, and report the maximum of two the metric values. We only include trainings in this analysis where the network reached 99\% training accuracy in predicting the final GoL state. We measure the mean metric value and impact of the following hyper-parameters in table \ref{tab:timestep-results}:
\begin{itemize}
\item The number of timesteps in GoL and model (\textit{Num timesteps})
\item Whether the network was trained with an autoencoder loss (\textit{Use Autoencoder})
\item Whether corresponding layers of different timesteps share the weights (\textit{Use RNN})
\end{itemize}

\begin{table}[ht]
\centering
\begin{tabular}[t]{c|c|c|c|c}
\hline
{} Num timesteps & Use Autoencoder & Use RNN & Number of Runs & Mean Metric Value \\
\hline
2 & True & True & 6 & 1.00 \\
3 & True & True & 42 & 0.99 \\
4 & True & True & 33 & 0.93 \\
3 & False & True & 22 & 0.56 \\
3 & True & False & 20 & 0.40 \\
\hline
\end{tabular}
\caption{\label{tab:timestep-results} The first 3 rows show that interpretability gets slightly worse with more timesteps. The last two rows show that the autoencoder and RNN are helpful. We average over \textit{Number of runs} different training runs of the network.}
\end{table}

We also do an evaluation where we train 15 versions of the network with 3 timesteps and measure how many training steps it takes each predicted state to reach 98\% similarity with the corresponding ground truth state (the states should correspond since the number of timesteps are equal). On average, the 3rd state reaches 98\% accuracy 2773 training steps before the 4th state. The 2nd state reaches 98\% accuracy 5533 training steps before the 4th state. The network is learning to predict the intermediate states before the final one, even though it was trained to predict the final state, but not the intermediate ones. We think the network is learning to predict the intermediate states first because they are necessary for predicting the final state.

We also measure the impact of other hyperparameters. We find that having 20 channels in our CNN layers improves the metric by 0.09 over larger numbers of channels. We also find that having 1 layer in the encoder and decoder improves the metric by 0.14 compared to 3 layers.

\subsection{Results with a different number of model and GoL timesteps}

We also measure the results when the number of model and GoL timesteps are different as shown in table \ref{tab:different-timesteps-results}. This is a more difficult problem because the model cannot simply use each model timestep to predict one GoL timestep.

\begin{table}[ht]
\centering
\begin{tabular}[t]{c|c|c|c}
\hline
{} GoL timesteps & Model timesteps &  Number of runs &  Mean metric value \\
\hline
2 & 3 & 16 & 0.73\\
3 & 2 & 22 & 0.74\\
\hline
\end{tabular}
\caption{\label{tab:different-timesteps-results}Results where the number of GoL and model timesteps are different. Note that the runs with 2 GoL timesteps and 3 model timesteps can have a metric value of 1.5 if both of their intermediate model states match the single intermediate ground truth state. These runs use the RNN and the autoencoder.}
\end{table}

\subsection{Impact of adversarial training}

\begin{table}[ht]
\begin{tabular}[t]{p{15mm}|p{15mm}|p{20mm}|p{10mm}|p{10mm}|p{20mm}|p{20mm}}
\hline
{} GoL timesteps & Model timesteps & Use auto-encoder & Use RNN &  Number of runs &  Mean without adversarial  &  Mean with adversarial \\
\hline
2 & 3 & True & True & 16 &  0.58 & 0.65 \\
3 & 2 & True & True & 22 &  0.15 & 0.71 \\
3 & 3 & True & False & 20 &  0.11 & 0.40 \\
3 & 3 & False & True & 10 &  0.44 & 0.43 \\
\hline
\end{tabular}
\caption{\label{tab:adver-results}Results on a subset of experiments with and without the adversarial training technique. The adversarial training technique improves the results in every case except when we do not use the autoencoder.}
\end{table}

We measure the impact of our adversarial training technique and show it in table \ref{tab:adver-results}. The technique usually improves results, sometimes by about 4x. Adversarial training helps more in the cases where the interpretability score is low without it. 

\section{Related work} \label{relatedwork}
\cite{logit_lens} also applies the decoder to intermediate layers to help interpretability. The difference with our work is that they apply the technique to a different architecture and do not have the goal of visualizing intermediate environment states. They also do not use an adversarial technique or have results showing how architecture decisions impact interpretability. See appendix \ref{appen:related_work} for more related work.

\section{Conclusion and future work}
In our research, we explore a new interpretability problem of visualizing the hidden intermediate states that a network represents. We show that in a simple environment using an autoencoder and adversarial techniques, we can obtain hidden states that closely match the ground truth states of the environment. However, the idea didn't end up working out when we tried it on a larger neural network trained on chess, highlighting the limitations of our approach for more complex models and domains. In future work, we plan to investigate ways to extend our technique to handle such complex scenarios, where the network may represent the consequences of a decision it is considering.

\bibliography{iclr2021_conference}
\bibliographystyle{iclr2021_conference}
\newpage
\appendix
\section{More related work \label{appen:related_work}}
 NNs have produced unprecedented advances in recent decades in a variety of tasks. However, their inner workings are still unclear. A large number of previous works have focused on visualizing NNs outputs, activation function and hidden layers. For example, \cite{Yosinskivisnn} provides a tool to visualize neurons in pre-trained Convolutional NNs (CNNs). Their method allows for deeper local understanding of neuron computations. This tool  only works for images and videos. 

Others such as \cite{intriguingnns} show that CNNs encoding maintains photographically accurate information about the original input several layers in the network. Their approach uses shallow and deep representation to invert an objective function with Stochastic Gradient Descent (SGD) that allows them to reconstruct the original input. Our method also sheds light on what the intermediate layers of a network represent. 


Another popular approach for unpacking the inner-workings of NNs has been via visualization. More specifically, the use of saliency maps \citep{saliencymaps} highlights an area (map) of the image such that the classifier (e.g. CNN) can discriminate given a class. This method has proved to be a powerful tool for image region segmentation. 


More recently, \cite{mordvintsev2020growing} investigated learning a cellular automata update rule. They use a recurrent network with simplified CNN-like layers, with ‘per-pixel’ dropout. They train it on the task of generating target images. Their visualizations show that the network also learns interpretable intermediate representations. One difference with our method is that we use a standard CNN architecture instead of a modified one. We also use a decoder which we apply to the intermediate representations, where they treat the last 3 channels as the visualization instead of using a decoder. We also apply our technique to a domain with complex intermediate states, while their intermediate states are usually subsets of the target image. We plan to further test our technique by applying it in the domain of generating target images.

Finally, the authors of \cite{viscnn} consider the task of visualizing the intermediate feature layers of a CNN as well as the operation used for classification. Their approach shows significant improvement on ImageNet. Our work differs in that instead of trying to extract specific features from the hidden states, we focus on reconstructing environment states.



\newpage
\section{Results samples \label{appen:results-samples}}
These are cherry picked examples to illustrate how our metric works.
\begin{figure}[h!]
  \centering
  \includegraphics[width=.75\columnwidth]{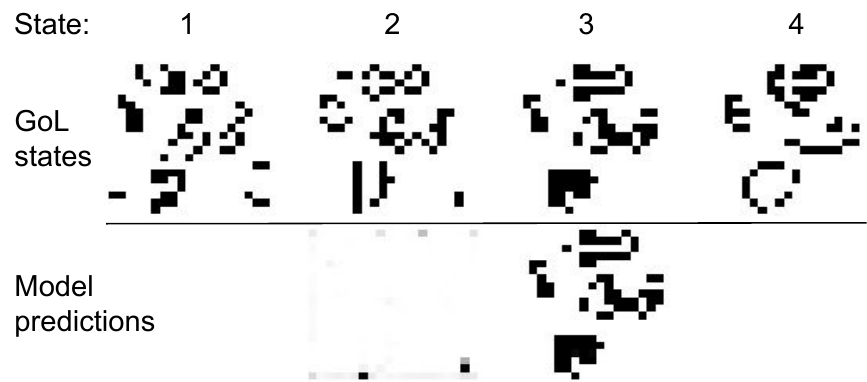}
  \caption{An example with 3 GoL timesteps and 3 model timesteps. The metric gets a value of .5 because the 3rd predicted state matches the ground truth, but the 2nd predicted state does not. The 1st and 4th states are ignored because the network is trained on them.}
\end{figure}

\begin{figure}[h!]
  \centering
  \includegraphics[width=.75\columnwidth]{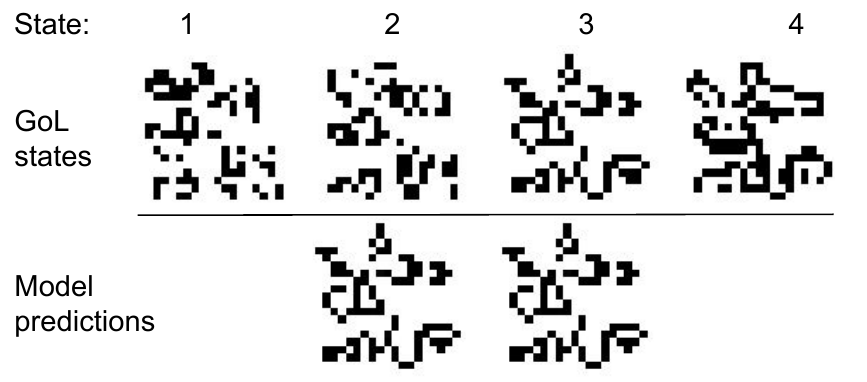}
  \caption{The metric gets a value of .75 because both predicted states match the same ground truth state. So the 3rd predicted state gets a half score.}
\end{figure}

\begin{figure}[h!]
  \centering
  \includegraphics[width=.75\columnwidth]{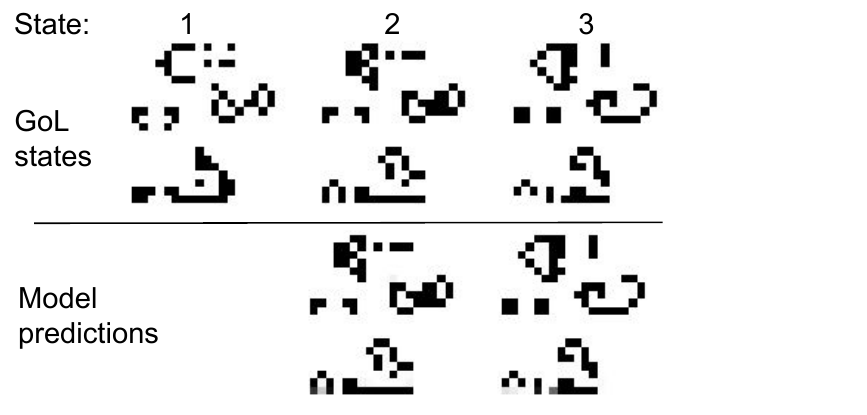}
  \caption{An example with 2 GoL timesteps and 3 model timesteps. The metric gets a value of 1.0. The 2nd model state matches the intermediate GoL state. The 3rd model state matches the 3rd GoL state, but this is ignored because the model was trained to predict the 3rd GoL state. The denominator in the metric is 1 because $min(2, 3-2) = 1$}
\end{figure}

\begin{figure}[h!]
  \centering
  \includegraphics[width=.75\columnwidth]{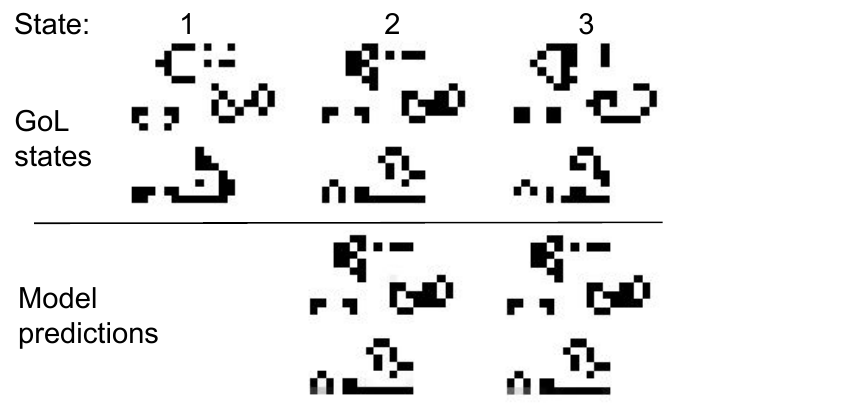}
  \caption{An example with 2 GoL timesteps and 3 model timesteps. The metric gets a value of 1.5. The 2nd model state matches the intermediate GoL state. The 3rd model state also matches the intermediate GoL state, so the metric gives .5 score for this state. Note that we ignore the last GoL state and not the last model state, to this match is a valid one}
\end{figure}

\end{document}